\documentclass[10pt,twocolumn,letterpaper]{article}

\usepackage[table]{xcolor}
\usepackage{iccv}
\usepackage{times}
\usepackage{epsfig}
\usepackage{graphicx}
\usepackage{amsmath, bm}
\usepackage{amssymb}
\usepackage{siunitx}
\usepackage{multicol}
\usepackage{multirow}
\usepackage{booktabs}
\usepackage{cuted}
\usepackage{capt-of}
\usepackage[pagebackref=true,breaklinks=true,letterpaper=true,colorlinks,bookmarks=false]{hyperref}

\iccvfinalcopy

\def\iccvPaperID{6166}

\newcommand{\comment}[1]{}

\newcommand{\Modelname}{H3D-Net} 

\newcommand{\bc}{\mathbf{c}}

\newcommand{\bn}{\mathbf{n}}

\newcommand{\br}{\mathbf{r}}

\newcommand{\bv}{\mathbf{v}}

\newcommand{\bx}{\mathbf{x}}

\newcommand{\bz}{\mathbf{z}}

\newcommand{\bC}{\mathbf{C}}

\newcommand{\bI}{\mathbf{I}}

\newcommand{\bM}{\mathbf{M}}

\newcommand{\mF}{\mathcal{F}}
\newcommand{\mG}{\mathcal{G}}

\newcommand{\mL}{\mathcal{L}}

\newcommand{\mbR}{\mathbb{R}} 

\newcommand{\argmin}{\operatornamewithlimits{arg\,min}}

\newcommand{\norm}[1]{\left\lVert#1\right\rVert}

\ificcvfinal\fi

\begin{document}
\title{\Modelname: Few-Shot High-Fidelity 3D Head Reconstruction}
\author{Eduard Ramon$^{1,2}$
\and
Gil Triginer$^1$
\and
Janna Escur$^1$
\and
Albert Pumarola$^3$
\and
Jaime Garcia$^{1}$
\and
Xavier Giro-i-Nieto$^{2,3}$
\and
Francesc Moreno-Noguer$^{3}$
\and
\small$^1$\emph{Crisalix SA}
\quad \small$^2$\emph{Universitat Polit\`{e}cnica de Catalunya}
\quad \small$^3$\emph{Institut de Robòtica i Informàtica Industrial, CSIC-UPC}\\
\url{crisalixsa.github.io/h3d-net}
\vspace{-2mm}
}

\maketitle
\ificcvfinal\thispagestyle{empty}\fi

\begin{strip}\centering
\vspace{-3em}
\includegraphics[width=\textwidth]{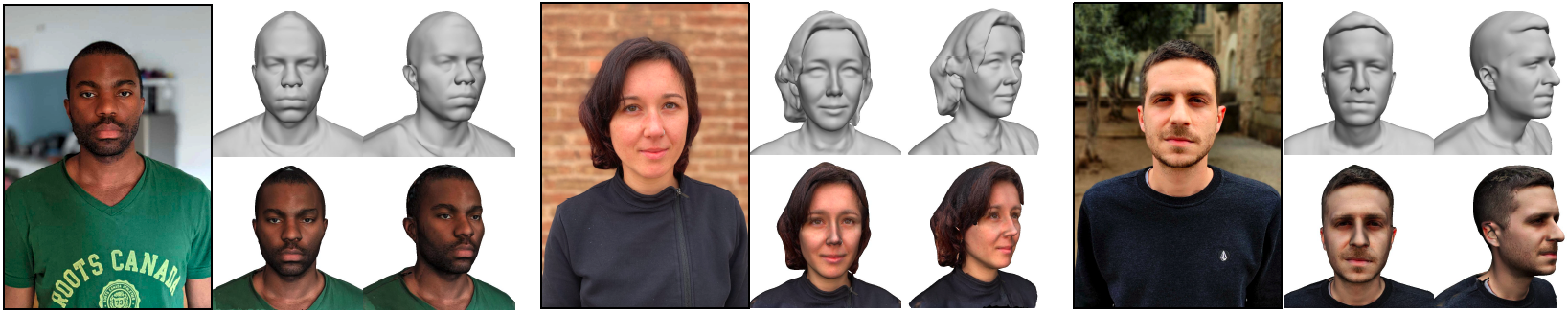}
\captionof{figure}{We introduce H3D-Net, a method for high-fidelity 3D head reconstruction in the wild. Our method estimates a signed distance function (SDF) of the head by optimizing a coordinate-based neural network on a small set of input images. This optimization process is constrained by a pre-trained probabilistic model of 3D head SDFs to obtain plausible shapes in few-shot setups. The figure shows the 3D head reconstruction of three scenes obtained with the proposed method from only three images with associated masks and camera poses.}\label{fig:teaser}
\end{strip}

\begin{abstract}
Recent learning approaches that implicitly represent surface geometry using coordinate-based neural representations have shown impressive results in the problem of multi-view 3D reconstruction. The effectiveness of these techniques is, however, subject to the availability of a large number (several tens) of input views of the scene, and computationally demanding optimizations. In this paper, we tackle these limitations for the specific problem of few-shot full 3D head reconstruction, by endowing coordinate-based representations with a probabilistic shape prior that enables faster convergence and better generalization when using few input images (down to three). First, we learn a shape model of 3D heads from thousands of incomplete raw scans using implicit representations. At test time, we jointly overfit two coordinate-based neural networks to the scene, one modeling the geometry and another estimating the surface radiance, using implicit differentiable rendering. We devise a two-stage optimization strategy in which the learned prior is used to initialize and constrain the geometry during an initial optimization phase. Then, the prior is unfrozen and fine-tuned to the scene. By doing this, we achieve high-fidelity head reconstructions, including hair and shoulders, and with a high level of detail that consistently outperforms both state-of-the-art 3D Morphable Models methods in the few-shot scenario, and non-parametric methods when large sets of views are available.
\vspace{-4mm}
\end{abstract}
\section{Introduction}
Recent learning based methods have shown impressive results in  reconstructing 3D shapes from 2D images. These approaches can be roughly split into two  main categories: model-based~\cite{bai2020deep, dou2018multi, Moreno_pami2013, Pumarola2018geometry, ramon2019multi, richardson20163d, richardson2017learning, tewari2017mofa, tran2018extreme, tuan2017regressing, wu2019mvf} and model-free~\cite{Cheng_2020_ACCV, choy20163d, jackson2017large, kar2017learning, Lin_2020_CVPR, pumarola2020c, pumarola20193dpeople, sela2017unrestricted, wang2018pixel2mesh, wei20193d, yan2016perspective}. The former incorporate prior knowledge obtained from training data to limit the space of feasible solutions, making these approaches  well suited for few-shot and one-shot shape estimation.  However, most model-based methods produce  shapes that usually lack geometric detail and cannot handle arbitrary topology changes.  

On the other hand, model-free approaches based on discrete representations like \eg voxels, meshes or point-clouds, have the flexibility to represent a wider spectrum of shapes, although at the cost of being computationally tractable only for small resolutions or being restricted to fixed topologies. These limitations have been overcome by neural implicit representations~\cite{chen2019learning, mescheder2019occupancy, park2019deepsdf,  pumarola2020d, saito2019pifu}, which can represent both geometry and  appearance as a continuum, encoded in the weights of a neural network. \cite{niemeyer2020differentiable,yariv2020multiview} have shown the success of such representations in learning detail-rich 3D geometry directly from images, with no 3D ground truth supervision. Unfortunately, the performance of these methods is currently conditioned to the availability of a  large number of input views, which leads to a time consuming inference.

In this work we introduce \Modelname{}, a hybrid scheme that combines the strengths of model-based and model-free representations by incorporating prior knowledge into neural implicit models for category-specific multi-view reconstruction. We apply this approach to the  problem of few-shot  full  head reconstruction. In order to build the prior,  we first use  several thousands of raw incomplete scans to learn a space of Signed Distance Functions (SDF) representing 3D head shapes \cite{park2019deepsdf}. At inference, this learnt shape prior is used to initalize and guide the optimization of an Implicit Differentiable Renderer (IDR) \cite{yariv2020multiview} that, given a potentially  reduced number of input images, estimates the full head geometry. The use of the learned prior enables faster convergence during optimization and prevents it from being trapped into local minima, yielding 3D shape estimates that capture fine details of the face, head and hair from just three input images (see. Fig.~\ref{fig:teaser}).

We exhaustively evaluate our approach on a mid-resolution Multiview-Stereo (MVS) public dataset \cite{pillai20192nd} and on a high-resolution dataset we collected with a structured-light scanner, consisting of 10 3D full-head scans. The results show that we consistently outperform current state-of-the-art, both in a few-shot setting and when many input views are available. Importantly, the use of the prior also  makes our approach very efficient, achieving competitive results in terms of accuracy about 20$\times$ faster than IDR~\cite{yariv2020multiview}. Our key contributions can be summarized as follows:
\begin{itemize}
\setlength{\itemsep}{0pt}
    \item We introduce a method for reconstructing high quality full heads in 3D from small sets of in-the-wild images.
    \item Our method is the first to use implicit functions for reconstructing 3D humans heads from multiple images and also to rival parametric and non-parametric models in 3D accuracy at the same time.
    \item We devise a guided optimization approach to introduce a probabilistic shape prior into neural implicit models. 
    \item We collect and will release a new dataset containing high-resolution 3D full head scans, images, masks and camera poses for evaluation purposes, which we dub H3DS. 
\end{itemize}
\section{Related work}

\noindent\textbf{Model-based}. 3D Morphable Models (3DMMs)~\cite{booth20173d, booth20163d, li2017learning, paysan20093d, ploumpis2019combining, tran2019towards, tran2018nonlinear, tran2019learning} have become the \textit{de facto} representation used for few-shot 3D face reconstruction in-the-wild given that they lead to light-weight, fast and robust systems. Adopting 3DMMs as a representation, the 3D reconstruction problem boils down to estimating the small set of parameters that best represent a target 3D shape. This makes it possible to obtain 3D reconstructions from very few images~\cite{bai2020deep, dou2018multi, ramon2019multi, wu2019mvf} and even a single input
~\cite{richardson20163d, richardson2017learning, tewari2017mofa, tran2018extreme, tuan2017regressing}. Nevertheless, one of the main limitations of morphable models is their lack of expressiveness, specially for high frequencies. This issue has been addressed by learning a post processing that transfers the fine details from the image domain to the 3D geometry~\cite{Lin_2020_CVPR, richardson2017learning, tran2018extreme}. Another limitation of 3DMMs is their inability to represent arbitrary shapes and topologies. Thus, they are not suitable for reconstructing full heads with hair, beard, facial accessories and upper body clothing.

\vspace{1mm}
\noindent\textbf{Model-free}. Model-free approaches build upon more generic representations, such as voxel-grids or meshes, in order to gain expressiveness and flexibility. Voxel-grids have been extensively used for 3D reconstruction~\cite{choy20163d,  hane2017hierarchical, jackson2017large, kar2017learning, yan2016perspective} and concretely for 3D face reconstruction \cite{jackson2017large}. Their main limitation is that memory requirements grow cubically with resolution, and octrees~\cite{hane2017hierarchical} have been proposed to address this issue. On the other hand, meshes~\cite{Cheng_2020_ACCV, kanazawa2018learning, Lin_2020_CVPR, wang2018pixel2mesh} are a more efficient representation for surfaces than voxel-grids, and are suitable for graphics applications. Meshes have been proposed for 3D face reconstruction~\cite{Cheng_2020_ACCV, Lin_2020_CVPR} in combination with graph neural networks~\cite{bronstein2017geometric}. However, similarly to 3DMMs, meshes are also usually restricted to fixed topologies and are not suitable for reconstructing other elements beyond the face itself.

\vspace{1mm}
\noindent\textbf{Implicit representations}. Recently, implicit representations~\cite{littwin2019deep, mescheder2019occupancy, park2019deepsdf} have been proposed to jointly address the memory limitations of voxel grids and the topological rigidity of meshes. Implicit representations model surfaces as a level-set  of a coordinate-based continuous function, \eg a signed distance function or an occupancy function. These functions, usually implemented as multi-layer perceptrons (MLPs), can theoretically express any shape with infinite resolution and a fixed memory footprint. Implicit methods can be divided in those that, at inference time, perform a single forward pass of a previously trained model ~\cite{chen2019learning, mescheder2019occupancy, saito2019pifu}, and those that overfit a model to a set of input images through an optimization process using implicit differentiable rendering ~\cite{niemeyer2020differentiable, yariv2020multiview}. In the later, given that the inference is an optimization process, the obtained 3D reconstructions are more accurate. However, they are slow and require an important number of multi-view images, failing in few-shot setups as those we consider in this work.

\begin{figure*}[!t]
    \centering
    \includegraphics[width=\textwidth]{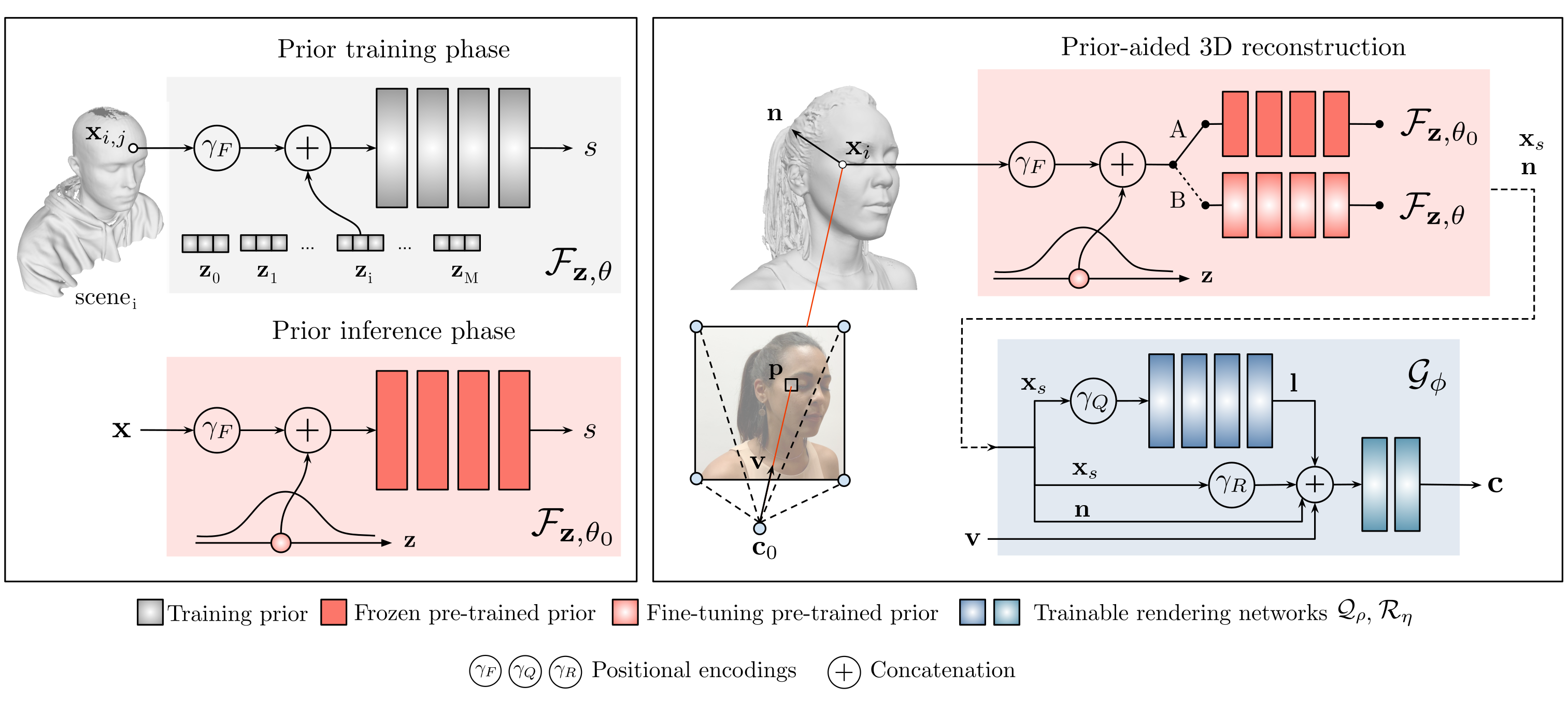}
    \caption{Overview of our method. {\bfseries Left.} The two configurations of the prior model at training and inference phases. {\bfseries Right.} Integration of the pre-trained prior model with the implicit differentiable renderer. During the prior-aided 3D reconstruction process, the geometry network starts off with frozen weights (commuter at position A), constraining the predicted shape to lie within its pre-learnt latent space, and is eventually unfrozen (commuter at position B) to allow fine-tuning of the fine details.}
    \label{fig:method}
\end{figure*}

\vspace{1mm}
\noindent\textbf{Priors for implicit representations}. Building priors for implicit representations has been addressed with two main purposes. The first consists in speeding up convergence of methods that perform an optimization at inference time~\cite{sitzmann2020metasdf} using meta-learning techniques~\cite{finn2017model}. The second is to find a space of implicit functions that represent a certain category using auto-decoders~\cite{park2019deepsdf, yenamandra2020i3dmm}. However, ~\cite{park2019deepsdf, yenamandra2020i3dmm} have been used to solve tasks using 3D supervision, and it is still an open problem how to use these priors when the supervision signal is generated from 2D images.

As done with morphable models, implicit shape models can be used to constrain image-based 3D reconstruction systems to make them more reliable. Drawing inspiration from this idea, in this work we leverage implicit shape models~\cite{park2019deepsdf} to guide the optimization-based implicit 3D reconstruction method~\cite{yariv2020multiview} towards more accurate and robust solutions, even under few-shot in-the-wild scenarios.
\section{Method}\label{sec:Method}

Given a small set of $N\geq3$ input images $\bI_v$, $v=1,\ldots,N$, with associated head masks $\bM_v$ and camera parameters $\bC_v$, our goal is to recover the 3D head surface $\mathcal{S}$ using only visual cues as supervision. Formally, we aim to approximate the signed distance function (SDF) $\mF:\bx \rightarrow s$ such that $\mathcal{S}$ $= \{\bx \in \mbR^3 | \mF(\bx)=0\}$.

In order to approximate $\mF$, we propose to optimize a previously learnt probabilistic model $\mF_{\bz, \theta_0}$, that represents a prior distribution over 3D head SDFs. $\bz$ and $\theta_0$ are a latent vector encoding specific shapes and the learnt parameters of an auto-decoder \cite{tan1995reducing}, respectively. Building on DeepSDF \cite{park2019deepsdf}, we learn these parameters from thousands of incomplete scans. We describe this process in section~\ref{ssec:Shape space}.

At test time, the reconstruction process is reduced to finding the optimal parameters $\{\bz^*,\theta^*\}$ such that $\mF_{\bz^*, \theta^*} \sim {\mF}$. To that end, we compose the prior model $\mF_{\bz,
\theta_0}$, which we also refer to as geometry network, with a rendering network $\mG_\phi: (\bx, \bn, \bv) \rightarrow \bc$ that models the RGB radiance emitted from a surface point $\bx$ with normal $\bn$ in a viewing direction $\bv$, and minimize a photometric error w.r.t. the input images $\bI_v$, as in \cite{yariv2020multiview}. Moreover, we propose a two-step optimization schedule that prevents the reconstruction process from getting trapped into local minima and, as we shall see in the results section, leads to much more accurate, robust and realistic reconstructions. We describe the reconstruction step in section \ref{ssec:3D Reconstruction}.

\subsection{Learning a prior for human head SDFs}
\label{ssec:Shape space}

Given a set of $M$ scenes with associated raw 3D point clouds, we use the DeepSDF framework to learn a prior distribution of signed distance functions representing 3D heads, $\mF_{\bz, \theta_0}$. While the original DeepSDF formulation requires watertight meshes as training data to use signed distances as supervision, we use the Eikonal loss \cite{gropp2020implicit} to learn directly from raw, and potentially incomplete, surface point clouds. In addition, Fourier features are used to overcome the spectral bias of MLPs towards low frequencies in low dimensional tasks~\cite{tancik2020fourier}. We illustrate the training and inference process of the prior model in figure~\ref{fig:method}-left. 

For each scene, indexed by $i=1,\ldots,M$, we sample a subset of points $\mathcal{P}_{\rm s}^{(i)}$ on the surface, and another set $\mathcal{P}_{\rm v}^{(i)}$ uniformly taken from a volume containing the scene, and minimize the following objective:
\begin{equation} \label{eq:deep_sdf_objective}
\argmin_{\{\bz_i\}, \theta} \sum_{i=1}^M \mL_{\rm Surf}^{(i)} + \lambda_0\mL_{\rm Emb}^{(i)} + \lambda_1\mL_{\rm Eik}^{(i)},
\end{equation}
where $\lambda_0$ and $\lambda_1$ are hyperparameters and $\mL_{\rm Surf}^{(i)}$ accounts for the SDF error at surface points:
\begin{equation} \label{eq:deep_sdf surface loss}
\mL_{\rm Surf}^{(i)} = \sum_{\bx_j \in \mathcal{P}_{\rm s}^{(i)}} |\mF_{\bz_i, \theta}(\bx_j)|\;.
\end{equation}

$\mL_{\rm Emb}^{(i)}$ enforces a zero-mean multivariate-Gaussian distribution
with spherical covariance $\sigma^2$ over the space of latent  vectors:
\begin{equation} \label{eq:deepsdf embedding loss}
\mL_{\rm Emb}^{(i)} = \frac{1}{\sigma^{2}}||\bz_i ||_2^2.
\end{equation}

Finally, $\mL_{\rm Eik}^{(i)}$ regularizes $F_{\bz_i, \theta}$ with the Eikonal loss to ensure that it approximates a signed distance function by keeping its gradients close to unit norm:
\begin{equation}\label{eq:deepsdf eikonal loss}
    \mL_{\rm Eik}^{(i)} = \sum_{\bx_k \in \mathcal{P}_{\rm v}^{(i)} } \big (\norm{\nabla_{\bx} \mF_{\bz_i, \theta}(\bx_k)}-1\big )^2.
\end{equation}

This regularization across the whole volume is necessary given that our meshes are not watertight and only a subset of surface points is available as ground truth. \cite{gropp2020implicit}.

After training, we have obtained the parameters $\theta_0$ that represent a space of human head SDFs. We can now draw signed distance functions of heads from $F_{\bz, \theta_0}$ by sampling the latent space $\bz$. We use this pre-trained model as the prior for the 3D reconstruction schedule described in the following section.

\subsection{Prior-aided 3D Reconstruction}
\label{ssec:3D Reconstruction}

Given a new scene, for which no 3D information is provided at this point, we aim to approximate the SDF that implicitly encodes the surface of the head by only supervising in the image domain. To that end, we compose the previously learnt geometry probabilistic model $\mF_{\bz, \theta_0}$ with the rendering network $\mG_\phi$, and supervise on the photometric error to find the optimal parameters $\bz^*$, $\theta^*$ and $\phi^*$. The reconstruction process is illustrated in figure~\ref{fig:method}-right.

For every pixel coordinate $p$ of each input image $\bI_v$, we march a ray $\br = \{\bc_0 + t\bv | t \geq 0\}$,  where $\bc_0$ is the position of the associated camera $\bC_v$, and $\bv$ the viewing direction. The intersection point $\bx_i$ between the ray $\br$ and the surface $\mathcal{S}_{\bz,\theta} = \{\bx | \mF_{\bz, \theta}(\bx) = 0\}$ can be efficiently found using sphere tracing \cite{hart1996sphere}. This intersection point can be made differentiable w.r.t $\bz$ and $\theta$ without having to store the gradients corresponding to all the forward passes of the geometry network, as shown in~\cite{niemeyer2020differentiable} and generalized by~\cite{yariv2020multiview}. The following expression is exact in value and first derivatives:
\begin{equation}\label{eq:surface}
 \bx_s = \bx_{i} - \frac{\bv}{\nabla_\bx \mF_{\bz_k, \theta_k}(\bx_{i}) \cdot \bv} \mF_{\bz, \theta}(\bx_{i})\;.
\end{equation}

Here $\bz_k$ and $\theta_k$ denote the parameters of $\mF_{\bz, \theta}$ at iteration $k$, and $\bx_s$ represents the intersection point made differentiable w.r.t. the geometry network parameters.

Next, we evaluate the mapping $\mG_\phi$ at $\bx_s$, $\bn=\nabla_\bx \mF_{\bz, \theta}(\bx_{s})$ and $\bv$ to estimate the color $\bc$ for the pixel $p$ in the image $\bI_v$:
\begin{equation}\label{eq:color}
 \bc = \mG_{\phi}(\bx_s, \bn, \bv)\;.
\end{equation}

Finally, in order to optimize the surface parameters $\bz$ and $\theta$, and the rendering parameters $\phi$, we minimize the following loss \cite{yariv2020multiview}:
\begin{equation} \label{eq:idr_loss}
\mL = \sum_{v=1}^N \mL_{\rm RGB}^{(v)} + \beta_0\mL_{\rm Mask}^{(v)} + \beta_1\mL_{\rm Eik}^{(v)},
\end{equation}
where $\beta_0$ and $\beta_1$ are hyperparameters. We next describe each  component of this loss. Let $\mathcal{P}$ be a mini-batch of pixels from view $v$, $\mathcal{P}_{\rm RGB}$ the subset of pixels whose associated ray intersects $\mathcal{S}_{\bz,\theta}$ and which have a nonzero mask value, and $\mathcal{P}_{\rm Mask} = \mathcal{P} \setminus \mathcal{P}_{\rm RGB}$. The $\mL_{\rm RGB}^{(v)}$ is the photometric error, computed as:
\begin{equation} \label{eq:idr_rgb_loss}
\mL_{\rm RGB}^{(v)} = \frac{1}{|\mathcal{P}|} \sum_{p\in \mathcal{P}_{\rm RGB}} |\bI_v(p) - \bc_v(p)|\;.
\end{equation}

$\mL_{\rm Mask}^{(v)}$ accounts for silhouette errors:

\begin{equation} \label{eq:idr_mask_loss}
\mL_{\rm Mask}^{(v)} = \frac{1}{\alpha|\mathcal{P}|} \sum_{p \in \mathcal{P}_{\rm Mask}} {\rm CE}(\bM_v(p), s_{v, \alpha}(p))\;,
\end{equation}
where $s_{v,\alpha} = {\rm sigmoid}(-\alpha \min_{t \geq 0} \mF_{\bz,\theta}(\br_t))$ is the estimated silhouette, $\rm CE$ is the binary cross-entropy and $\alpha$ is a hyperparameter. Lastly, $\mL_{\rm Eik}$ encourages $\mF_{\bz, \theta}$ to approximate a signed distance function as in equation \ref{eq:deepsdf eikonal loss}.

Instead of jointly optimizing all the parameters $\{\bz, \theta, \phi\}$ to minimize $\mL$ we introduce a two-step optimization schedule which is more appropriate for auto-decoders like DeepSDF. We begin by initializing the geometry network $\mF_{\bz, \theta}$ with the previously learnt prior for human head SDFs, $\mF_{\bz, \theta_0}$, and a randomly sampled $\bz_0$ such that $\norm{\bz_0} < \epsilon$ to stay near the mean of the latent space. In a first phase, we only optimize $\bz$ and $\phi$ as $\argmin_{\bz,\phi} \mL$, which is equivalent to the standard auto-decoder inference. By doing so, the resulting surface $\mathcal{S}_{z^*, \theta}$ is forced to stay within the learnt distribution of 3D heads. Once the geometry and the radiance mappings have reached an equilibrium, \ie the optimization has converged, we unfreeze the decoder parameters $\theta$ to fine-tune the whole model as $\argmin_{\bz,\theta,\phi}\mL$.

In section~\ref{sec:Experiments}, we empirically prove that by using this optimization schedule instead of optimizing all the parameters at once, the obtained 3D reconstructions are more accurate and less prone to artifacts, specially in few-shot setups.
\section{Implementation details}

\begin{figure}[t!]
    \centering
    \includegraphics[width=\linewidth]{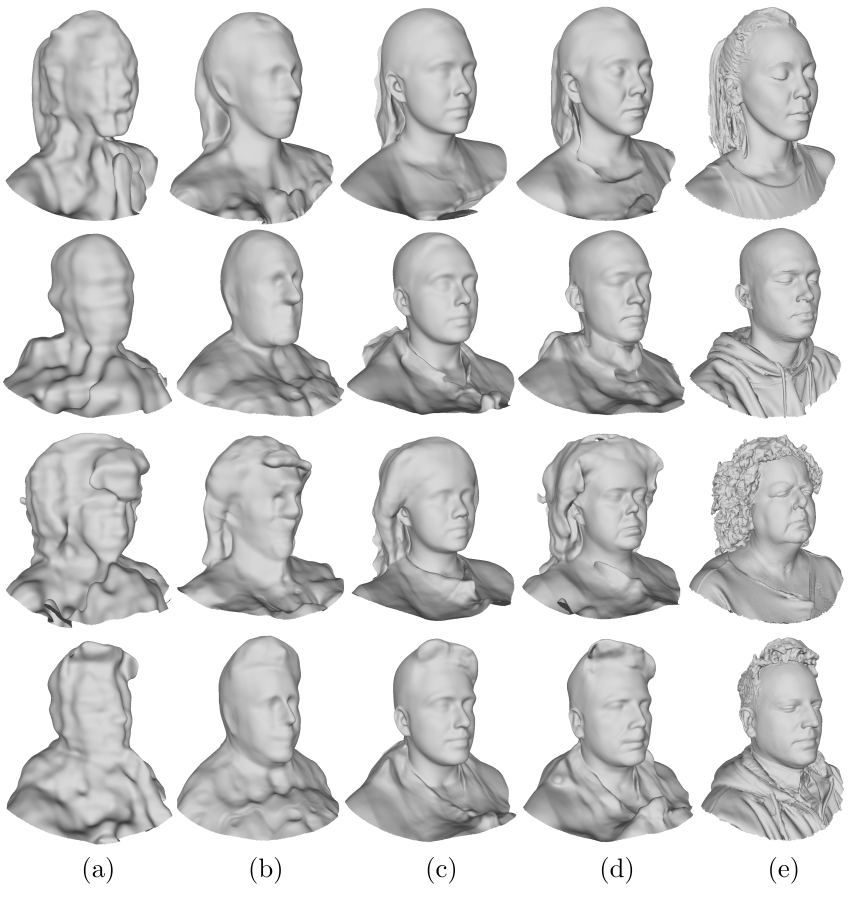}
    \caption{{\bfseries Ablation study} of our method in the few-shot setup (3 views). From left to right: (a) Ours with geometric initialization \cite{atzmon2020sal} and no schedule, (b) Ours with small prior initialization (500 subjects) and no schedule, (c) Ours with large prior initialization (10,000 subjects) and no schedule, (d) Ours with large prior initialization and schedule and (e) ground truth.}
    \label{fig:ablation}
\end{figure}

\begin{table}[t!]
\setlength{\tabcolsep}{4pt} 
\centering
\resizebox{0.5\textwidth}{!}{
\sisetup{detect-weight=true}
\begin{tabular}{ccccc}
\toprule
& (a) & (b) & (c) & (d) \\
\cmidrule{2-5}
Face mean distance [mm] & 4.04 & 2.68 & 1.90 & \bfseries 1.49  \\
Full-head mean distance [mm] & 16.68 & 17.08 & 14.59 & \bfseries 12.76 \\
\bottomrule
\end{tabular}
}
\vspace{2mm}
\caption{{\bfseries Ablation study} of our method in the few-shot setup (3 views). The face and full-head mean distances are the averages over all the subjects in the H3DS dataset. The configurations a,b,c,d are the same as those described in figure \ref{fig:ablation}.}
\label{table:ablation}
\vspace{-3mm}
\end{table}

Our implementation of the prior model closely follows the one proposed in \cite{gropp2020implicit}, with the addition that we apply a positional encoding $\gamma_\mathcal{F}$ to the input coordinates $\bx$ with 6 log-linear spaced frequencies. The encoded 3D coordinates are concatenated with the $\bz$ latent vector of size 256 and set as the input to the decoder. The decoder is a MLP of 8 layers with 512 neurons in each layer and single skip connection from the input of the decoder to the output of the 4th layer. We use Softplus as activation function in every layer except the last, where no activation is used. The prior model is trained for 100 epochs using Adam \cite{kingma2014adam} with standard parameters, learning rate of $10^{-4}$ and learning rate step decay of 0.5 every 15 epochs. The training takes approximately 50 minutes for a small dataset (500 scenes) and 10 hours for a large one (10,000 scenes).

\begin{figure}[t!]
    \centering
    \includegraphics[width=\linewidth]{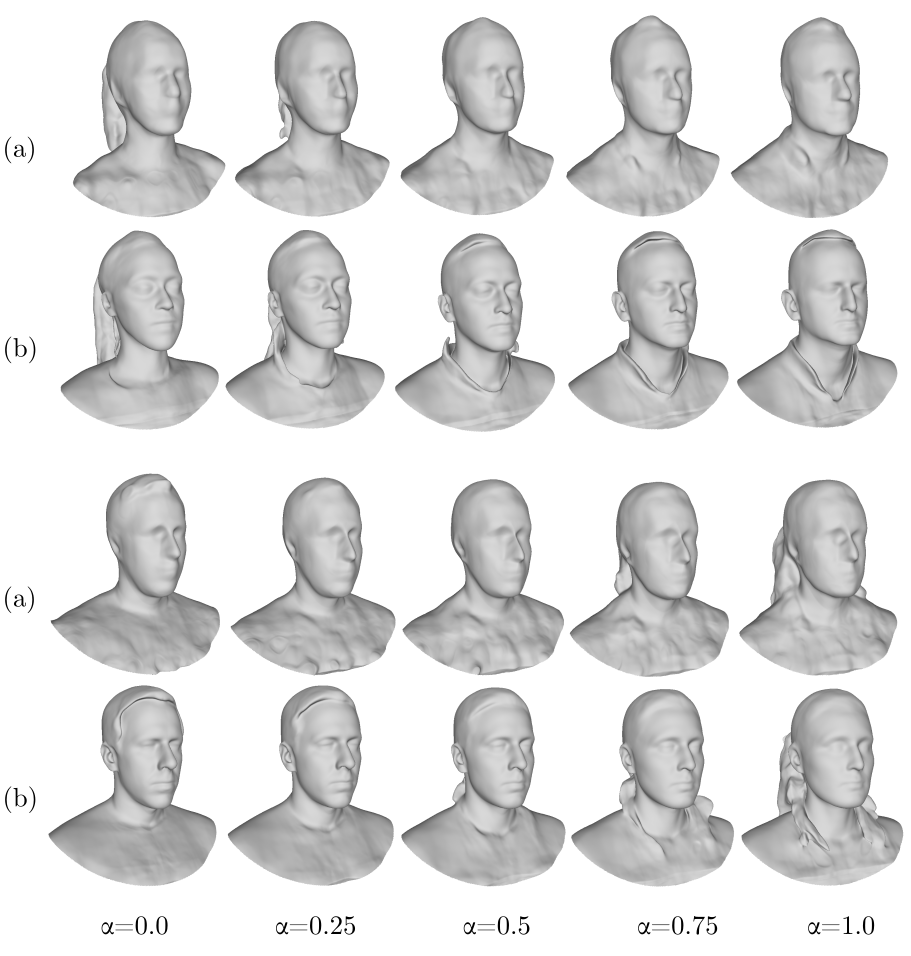}
    \caption{{\bfseries Latent interpolation} between different subjects, being $\alpha$ a linear interpolation factor in $\bz$ space. (a) uses the small prior model (500 subjects), and (b) the large prior (10,000 subjects).}
    \label{fig:latent}
\end{figure}

The 3D reconstruction network is composed by the prior model described above and a mapping $\mG_\phi$ that is split into two sub-networks $\mathcal{Q}_\rho$ and $\mathcal{R}_\eta$ as shown in figure \ref{fig:method}. $\mathcal{Q}_\rho$ is a MLP implemented exactly as the decoder of the prior model, except for the input layer, which takes in a 3 dimensional vector, and the output layer, which outputs a $d$-dimensional vector. As in \cite{yariv2020multiview}, $\mathcal{R}_\eta$ is a smaller MLP composed by 4 layers, each 512 neurons wide, no skip connections, and ReLU activations except in the output layer which is tanh. We also apply the positional encodings $\gamma_\mathcal{Q}$ and $\gamma_\mathcal{R}$ to $\bx_s$ with 6 and 4 log-linear spaced frequencies respectively.
Each scene is trained for 2000 epochs using Adam with fixed learning rate of $10^{-4}$ and learning rate step decay of 0.5 at epochs 1000 and 1500. The scene reconstruction process takes approximately 25 minutes for scenes of 3 views and 4 hours and 15 minutes for scenes of 32.

All the experiments for both prior and reconstruction models have been performed using a single Nvidia RTX 2080Ti.
\section{Experiments}\label{sec:Experiments}

In this section, we evaluate quantitatively and qualitatively our multi-view 3D reconstruction method. We empirically demonstrate that the proposed solution surpasses the state of the art in the few-shot \cite{bai2020deep, wu2019mvf} and many-shot \cite{yariv2020multiview} scenarios for 3D face and head reconstruction in-the-wild.

\definecolor{rowblue}{RGB}{220,230,240}

\begin{table*}[t!]
\setlength{\tabcolsep}{4pt} 
\centering
\rowcolors{7}{rowblue}{white}
\resizebox{0.75\textwidth}{!}{
\sisetup{detect-weight=true}
\begin{tabular}{lcccccccccccc}
\toprule
& \multicolumn{2}{c}{3DFAW} & \multicolumn{10}{c}{H3DS} \\
\cmidrule(lr{0.5em}){2-3} \cmidrule(lr{0.5em}){4-13}
& \multicolumn{2}{c}{3 views} & \multicolumn{2}{c}{3 views} & \multicolumn{2}{c}{4 views} & \multicolumn{2}{c}{8 views} & \multicolumn{2}{c}{16 views} & \multicolumn{2}{c}{32 views}  \\
\cmidrule{1-13}
& \multicolumn{2}{c}{face} & face & head & face & head & face & head & face & head & face & head \\
MVFNet \cite{wu2019mvf} & \multicolumn{2}{c}{1.54} & 1.66 & - & - &- & - & - & - & - & - & -\\
DFNRMVS \cite{bai2020deep} & \multicolumn{2}{c}{1.53}  & 1.83 & - & - &- & - & - & - & - & - & - \\
IDR \cite{yariv2020multiview} & \multicolumn{2}{c}{3.92} & 3.52 & 17.04 & 2.14 & 8.04 & 1.95 & 8.71 & 1.43 & 5.94 & 1.39 & 5.86 \\
\Modelname{} (Ours) & \multicolumn{2}{c}{\bfseries 1.37} & \bfseries 1.49 & \bfseries 12.76 & \bfseries 1.65 & \bfseries 7.95 & \bfseries 1.38 & \bfseries 5.47 & \bfseries 1.24 & \bfseries 4.80 & \bfseries 1.21 & \bfseries 4.90 \\
\bottomrule
\end{tabular}
}
\vspace{2mm}
\caption{{\bfseries 3D reconstruction method comparison.} Average surface error in millimeters computed over all the subjects in the 3DFAW and H3DS datasets. Find the precise definition of the face/head metrics, as well as a description of the distribution of the views, in section \ref{sec:experiments-setup}. }
\label{table:quantitative}
\vspace{-3mm}
\end{table*}

\begin{figure}[b!]
    \centering
    \includegraphics[width=\linewidth]{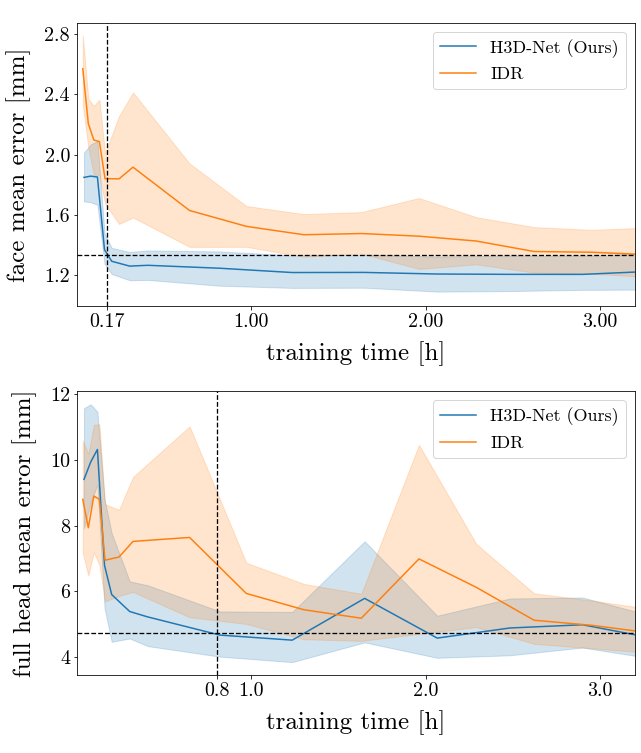}
    \caption{{\bfseries 3D reconstruction convergence} comparison between H3D-Net and IDR \cite{yariv2020multiview} using 32 views. Metrics are computed over all the samples in the H3DS dataset. The dotted lines indicate the time when our method first surpasses the best mean error attained by IDR over the entire optimization. {\bfseries Top.} Mean surface error in the face. {\bfseries Bottom.} Mean surface error in the full head.}
    \label{fig:convergence}
\end{figure}

\subsection{Datasets}

\vspace{1mm}
\noindent{\bf Prior training.} In order to train the geometry prior, we use an internal dataset made of 3D head scans from 10,000 individuals. The dataset is perfectly balanced in gender and diverse in  age and ethnicity. The raw data is automatically processed to remove internal mesh faces and non-human parts such as background walls. Finally, all the scenes are aligned by registering a template 3D model with non-rigid Iterative Closest Point (ICP) \cite{amberg2007optimal}.

\vspace{1mm}
\noindent{\bf 3DFAW \cite{pillai20192nd}.} We evaluate our method in the 3DFAW dataset. This dataset provides videos recorded in front, and around, the head of a person in static position as well as mid-resolution 3D ground truth of the facial region. We select 5 male and 5 female scenes and use them to evaluate only the facial region.

\vspace{1mm}
\noindent{\bf H3DS.} We introduce a new dataset called \textit{H3DS}, the first dataset containing high resolution full head 3D textured scans and 360º images with associated ground truth camera poses and ground truth masks. The 3D geometry has been captured using a structured light scanner, which leads to more precise ground truth geometries than the ones from 3DFAW \cite{pillai20192nd}, which were generated using Multi-View Stereo (MVS). The dataset consists of 10 individuals, 50\% man and 50\% woman. We use this dataset to evaluate the accuracy of the different methods in both the full head and the facial regions. We plan to release, maintain and progressively grow the H3DS dataset.

\subsection{Experiments setup}
\label{sec:experiments-setup}
We use the 3DMM-based methods MVFNet \cite{wu2019mvf} and DFNRMVS \cite{bai2020deep}, and the model-free method IDR \cite{yariv2020multiview} as baselines to compare against \Modelname{}.

In the few-shot scenario (3 views), all the methods are evaluated on the 3DFAW and H3DS datasets. To benchmark our method when more than 3 views are available, we compare it against IDR on the H3DS dataset.

\begin{figure*}[t!]
    \centering
    \includegraphics[width=1.\textwidth]{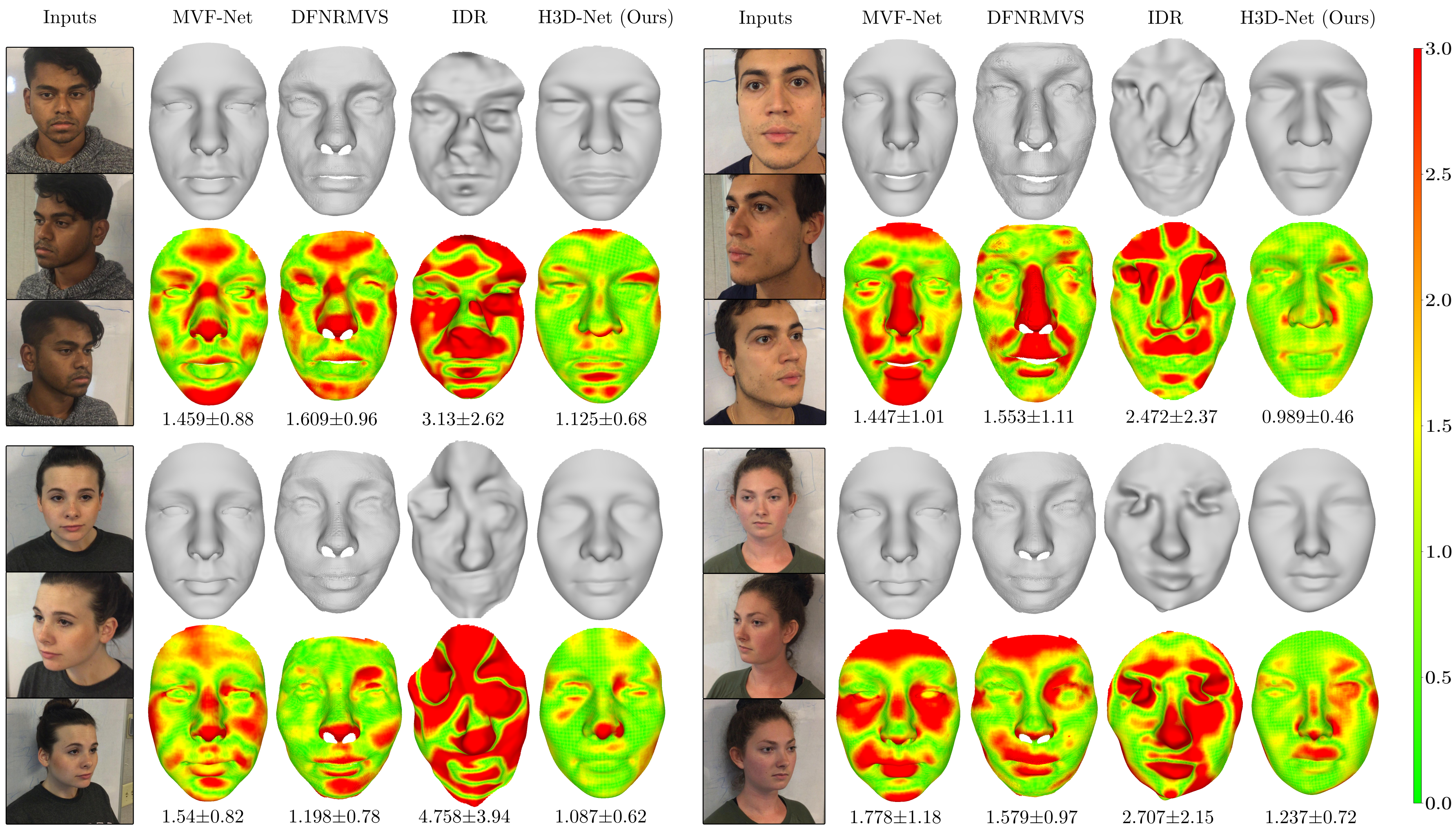}
    \caption{{\bfseries Qualitative results} obtained for 4 subjects from the 3DFAW dataset \cite{pillai20192nd} with only three input views. First and third rows show the reconstructed geometry and second and fourth rows show the surface error with the color code being in millimeters.}
    \label{fig:qualitative 3dmm}
\end{figure*}

The evaluation criteria have been the same for all methods and in all the experiments. The predicted 3D reconstruction is roughly aligned with the ground truth mesh using manually annotated landmarks, and then refined with rigid ICP \cite{besl1992method}. Then, we compute the unidirectional Chamfer distance from the predicted reconstruction to the ground truth. All the distances are computed in millimeters.

We report metrics in two different regions, the face and the full head. For the finer evaluation in the face region, we cut both the reconstructions and the ground truth using a sphere of 95 mm radius and with center at the tip of the nose of the ground truth mesh, and refine the alignment with ICP as in \cite{pillai20192nd, tuan2017regressing}. Then, we compute the Chamfer distance in this sub-region. For the full head evaluation, the ICP alignment is performed using an annotated region that includes the face, the ears, and the neck, since it is a region visible in all view configurations (3, 4, 8, 16 and 32). These configurations are defined by their yaw angles as follow: $\mathcal{V}_3=\{0, \pm45\}$, $\mathcal{V}_4=\{\pm45, \pm90\}$ and $\mathcal{V}_N = \{\frac{360}{N}i\}_{i=1}^N$ for $N = {8,16,32}$. In this case, the Chamfer distance is computed for all the vertices of the reconstruction.

\subsection{Ablation study}

We conduct an ablation study on the H3DS dataset in the few-shot scenario (3 views) and show the numerical results in table \ref{table:ablation}, and the qualitative results in figure \ref{fig:ablation}. First, we reconstruct the scenes without prior and without schedule (a). In this case, the geometry network is initialized using geometric initialization \cite{atzmon2020sal}, representing a sphere of radius one at the beginning of the optimization. Then, we initialize the geometry network with two different priors, a small one trained on 500 subjects (b), and a large one trained on 10,000 subjects (c), and perform the reconstructions without schedule. As it can be observed, initializing the geometry network with a previously learnt prior leads to smoother and more plausible surfaces, specially when more subjects have been used to train it. It is important to note that the benefits of the initialization are not only due to a better initial shape, but also to the ability of the initial weights to generalize to unseen shapes, which is greater in the large prior model. Finally, we initialize the geometry network with the large prior and use the proposed optimization schedule during the reconstruction process. It can be observed how the resulting 3D heads resemble much more to the ground truth in terms of finer facial details.

\begin{figure*}[t!]
    \centering
    \includegraphics[width=0.98\textwidth]{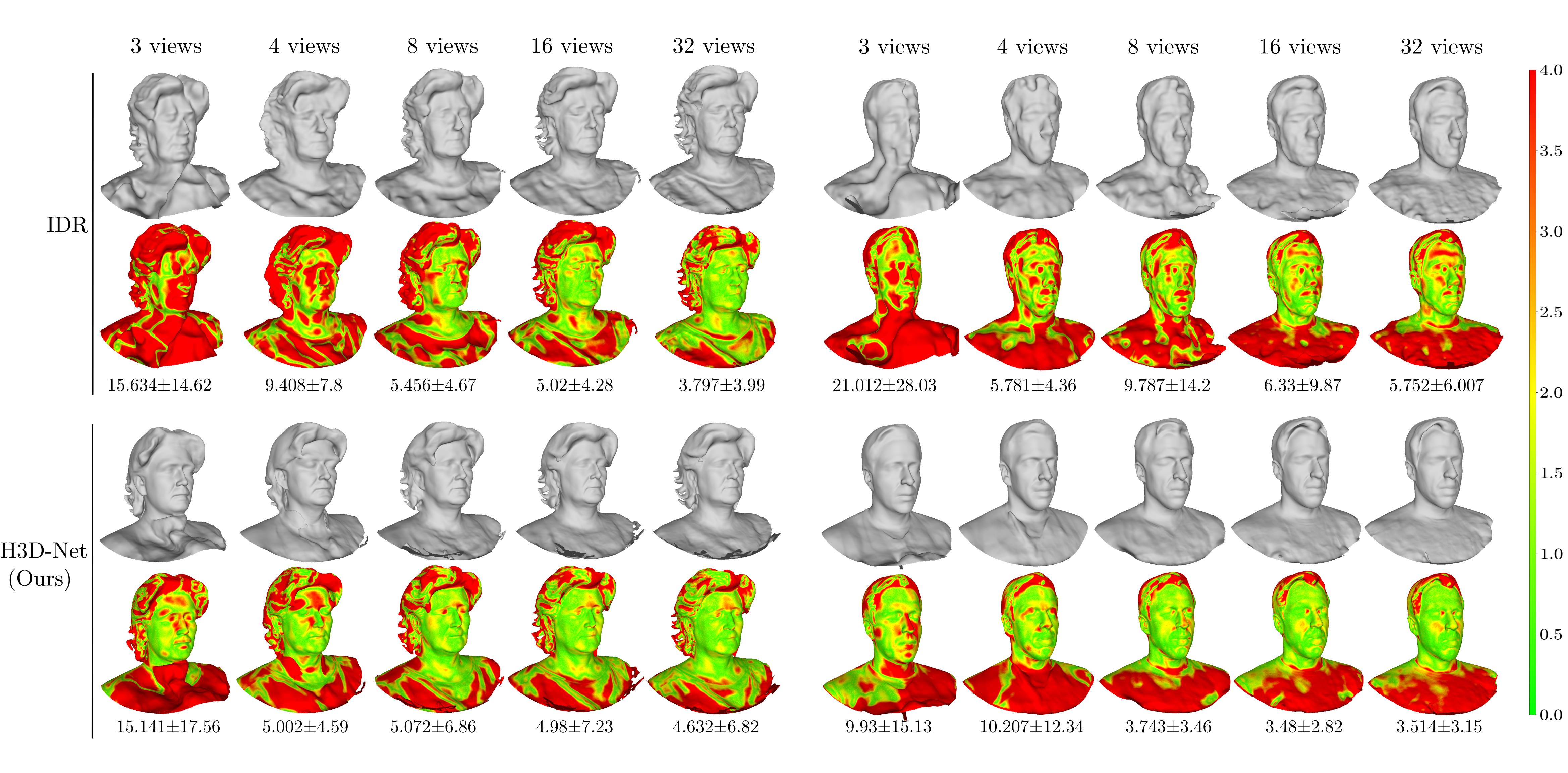}
    \caption{{\bfseries Qualitative results} obtained for 2 subjects from the H3DS dataset when varying the number of views. The first and second rows correspond with the results of IDR and the third and the forth with \Modelname{} results (ours). The surface error is represented with the color code in millimeters.}
    \label{fig:qualitative idr}
\end{figure*}

Given the notable effect that the number of samples has in the learnt prior representations and in the resulting 3D reconstructions as well, we visualize latent space interpolations in figure~\ref{fig:latent}. To that end, we optimize the latent vector for two ground truth 3D scans as shown in figure~\ref{fig:method}-left-bottom in order minimize the loss \ref{eq:deep_sdf_objective}. Then, we interpolate between the two optimal latent vectors. As it can be observed, the 3D reconstructions resulting from the interpolation in $\bz$ space of the large prior model are more detailed and plausible than the ones from the small prior model, suggesting that the later achieves poorer generalization.

\subsection{Quantitative results}

Quantitative results in terms of surface error are reported in table \ref{table:quantitative}. Remarkably, \Modelname{} outperforms both 3DMM-based methods in the few-shot regime, and the model-free method IDR when the largest number of views (32) are available. It is worth noting how the enhancement due to the prior is more significant as the number of views decreases. Nevertheless, the prior does not prevent the model from becoming more accurate when more views are available, which is a current limitation of model-based approaches.

We also analyze the trade-off between the optimization time and the accuracy in IDR and \Modelname{} for the case of 32 views, which we illustrate in figure \ref{fig:convergence}. It can be observed that, despite reaching similar errors asymptotically, in average our method achieves the best performance attained by IDR much faster. In particular, we report convergence gains of 20$\times$ for the facial region error and 4$\times$ for the full head. Moreover, the smaller variance observed in \Modelname{} (blue) indicates that it is a more stable method.

\subsection{Qualitative results}

Quantitative results show improvements over the baselines in both few-shot and many-shot setups. Here, we study how this is translated into the reconstructed 3D shape.

In figure \ref{fig:qualitative 3dmm}, we qualitatively evaluate the three baselines and \Modelname{} for the case of 3 input views. As expected, IDR \cite{yariv2020multiview} is the worst performing model in this scenario, generating reconstructions with artifacts and with no resemblance to human faces. On the other hand, 3DMM-based models \cite{bai2020deep, wu2019mvf} achieve more plausible shapes, but they struggle to capture fine details. \Modelname{}, in contrast, is able to capture much more detail and reduce significantly the errors over the whole face and, concretely, in difficult areas such as the nose, the eyebrows, the cheeks, and the chin.

We also evaluate the impact that varying the number of available views has
on the reconstructed surface, and compare our method to IDR \cite{yariv2020multiview}. As shown in figure \ref{fig:qualitative idr}, \Modelname{} is able to obtain surfaces with less error (greener) with far fewer views, which is consistent with the quantitative results reported in table \ref{table:quantitative}. Notably, it can also be observed that, even when errors are numerically similar (first and third columns), the reconstructions from \Modelname{} are much more realistic. In addition, \Modelname{} improvements are especially notable within the face region. We attribute this to the fact that training data used to build the prior model is more rich in this area, whereas training examples frequently present holes in other parts of the head.

\section{Conclusions}

In this work we have presented \Modelname, a method for high-fidelity 3D head reconstruction from small sets of in-the-wild images with associated head masks and camera poses. Our method combines a pre-trained probabilistic model, which represents a distribution of head SDFs, with an implicit differentiable renderer that allows direct supervision in the image domain. By constraining the reconstruction process with the prior model, we are able to robustly recover detailed 3D human heads, including hair and shoulders, from only three input images. After a thorough quantitative and qualitative evaluation, our experiments show that our method outperforms both model-based methods in the few-shot setup and model-free methods when a large number of views are available. One limitation of our method is that it still requires several minutes to generate 3D reconstructions. An interesting direction for future work is to use more efficient representations for SDFs in order to speed up the optimization process. We also find it promising to introduce texture priors, which could reduce the convergence time and reduce the final overall error.
\section{Acknowledgments}
This work has been partially funded by the Spanish government with the projects MoHuCo PID2020-120049RB-I00, DeeLight PID2020-117142GB-I00 and Maria de Maeztu Seal of Excellence MDM-2016-0656, and by the Government of Catalonia under the industrial doctorate 2017 DI 028.

{\small
\bibliographystyle{ieee_fullname}
\bibliography{main}
}

\end{document}